\documentclass{article}

\usepackage[preprint]{neurips_2024}

\usepackage[utf8]{inputenc} 
\usepackage[T1]{fontenc}    
\usepackage{hyperref}       
\usepackage{url}            
\usepackage{booktabs}       
\usepackage{amsfonts}       
\usepackage{nicefrac}       
\usepackage{microtype}      
\usepackage{xcolor}         
\usepackage{graphicx}       

\title{Mitigating Catastrophic Forgetting in Mathematical Reasoning Finetuning through Mixed Training}

\author{%
John Graham Reynolds \\
  Department of Computer Science \\
  The University of Texas at Austin \\
  Austin, TX 78712 \\
  \texttt{johngrahamreynolds@utexas.edu} \\
}

\begin{document}

\maketitle

\begin{abstract}
    When finetuning large language models for specialized tasks such as mathematical 
    reasoning, models exhibit catastrophic forgetting, losing previously learned 
    capabilities. We investigate this by finetuning Flan-T5-Base (250M parameters) on 
    the DeepMind Mathematics dataset and measuring forgetting on MultiNLI. Math-only 
    training improves mathematical accuracy from 3.1\% to 12.0\% but causes NLI accuracy 
    to collapse from 81.0\% to 16.5\%—a 64.5 percentage point drop occurring within the 
    first 1,000 training steps. We propose mixed training strategies that interleave 
    mathematical and NLI examples during training. Our results demonstrate that mixed 
    training completely eliminates catastrophic forgetting while maintaining equivalent 
    mathematical performance: the balanced 1:1 ratio achieves 12.0\% math accuracy 
    (matching math-only) while preserving 86.2\% NLI accuracy. We systematically explore 
    mixing ratios from 1:1 to 15:1, finding that even minimal NLI exposure (6.2\%) 
    provides effective regularization. These findings demonstrate that specialization 
    need not require forgetting general capabilities, with implications for scaling to 
    larger models where mixed training may confer additional benefits beyond forgetting 
    prevention.\footnote{Code, data, and trained models available at 
    \url{https://github.com/johngrahamreynolds/mathematical_catastrophe_mitigation} 
    and \url{https://huggingface.co/MarioBarbeque}.}
\end{abstract}

\section{Introduction}

Large language models (LLMs) pretrained on diverse corpora have demonstrated remarkable capabilities across a wide 
range of natural language understanding and generation tasks. However, when these models are finetuned for specialized 
domains such as mathematical reasoning, they often exhibit \emph{catastrophic forgetting}---a phenomenon where 
performance on previously learned tasks degrades significantly (\cite{kirkpatrick2017overcoming, french1999catastrophic}).

This forgetting poses a critical challenge for deploying specialized models in production settings where maintaining 
general language understanding capabilities is essential. For instance, a model finetuned for mathematical problem-solving 
should retain its ability to understand natural language, perform reasoning, and handle diverse linguistic phenomena.

In this work, we investigate catastrophic forgetting in the context of mathematical reasoning finetuning. Specifically, 
we finetune Flan-T5-Base (\cite{chung2022scaling}) on the DeepMind Mathematics dataset (\cite{saxton2019analyzing}) and measure 
the resulting degradation on the MultiNLI natural language understanding task (\cite{williams2018multinli}). Our experiments 
reveal severe catastrophic forgetting: math-only training improves mathematical reasoning accuracy from 3.1\% to 12.0\%, 
but causes NLI accuracy to drop dramatically from 81.0\% to 16.5\%.

To mitigate this forgetting, we propose and systematically evaluate \emph{mixed training} strategies that interleave 
mathematical and NLI examples during training. Our key contributions are:

\begin{itemize}
    \item We provide empirical evidence of severe catastrophic forgetting when finetuning for mathematical reasoning, 
    demonstrating a 64.5 percentage point drop in NLI performance.
    \item We show that balanced mixed training (1:1 ratio) completely eliminates 
    catastrophic forgetting while maintaining mathematical performance equivalent 
    to math-only training (12.0\% vs 12.0\%), achieving 86.2\% NLI accuracy. This demonstrates 
    that specialization need not require forgetting general capabilities.
    \item We systematically explore mixing ratios from 1:1 to 15:1, revealing a clear trade-off between mathematical 
    performance and NLI retention.
    \item We provide practical guidance for practitioners on selecting appropriate mixing ratios based on their 
    performance requirements.
\end{itemize}

\section{Related Work}

\subsection{Catastrophic Forgetting}

Catastrophic forgetting was first identified in neural networks by (\cite{mccloskey1989catastrophic}) and has been 
extensively studied in the context of continual learning (\cite{parisi2019continual, delange2021continual}). The 
phenomenon occurs when neural networks, optimized for a new task, lose previously acquired knowledge.

Several approaches have been proposed to mitigate catastrophic forgetting, including regularization-based methods 
(\cite{kirkpatrick2017overcoming, zenke2017continual}) and replay-based methods (\cite{rebuffi2017icarl}). However, most 
of this work has focused on computer vision tasks, with less attention paid to language models and specialized 
finetuning scenarios.

\subsection{Mathematical Reasoning in Language Models}

Recent work has demonstrated that language models can be finetuned for mathematical reasoning tasks 
(\cite{cobbe2021training, lewkowycz2022solving, hendrycks2021measuring}). The DeepMind Mathematics dataset 
(\cite{saxton2019analyzing}) has become a standard benchmark for evaluating mathematical reasoning capabilities. 
However, these studies have primarily focused on improving mathematical performance without considering the impact 
on general language understanding.

\subsection{Mixed Training and Multi-Task Learning}

Multi-task learning (\cite{caruana1997multitask}) has long been recognized as a way to improve generalization and 
prevent overfitting. In the context of language models, mixed training has been used in instruction tuning 
(\cite{chung2022scaling, wei2022finetuned}) to improve zero-shot generalization. However, the specific application 
of mixed training to mitigate catastrophic forgetting during specialized finetuning has received less attention.

\section{Methodology}

\subsection{Model and Datasets}

We use Flan-T5-Base (\cite{chung2022scaling}), a 250M parameter encoder-decoder language model that has been 
instruction-tuned on a diverse collection of tasks. This model provides a strong baseline for both natural language 
understanding and quantitative reasoning problems.

\textbf{Mathematical Reasoning Dataset:} We use the DeepMind Mathematics dataset (\cite{saxton2019analyzing}), specifically the linear algebra 1D subset. 
This dataset contains mathematical problems of the form ``Solve $24 = 1601c - 1605c$ for $c$.'' with numerical answers. 
We subsample the training set to 392,702 examples to match the size of our NLI dataset, ensuring fair comparison across experiments.

\textbf{Natural Language Understanding Dataset:} We use the MultiNLI (MNLI) dataset (\cite{williams2018multinli}), which 
consists of premise-hypothesis pairs labeled as entailment, contradiction, or neutral. The dataset contains 392,702 
training examples and 9,815 validation examples (matched split). We format examples as ``mnli premise: \{premise\} 
hypothesis: \{hypothesis\}'' to match Flan-T5's instruction format.

\subsection{Mixed Training Strategy}

Our mixed training approach interleaves batches from both datasets during training according to a specified ratio. For 
a mixing ratio of $m:n$ (math:NLI), each training step consists of $m$ math batches followed by $n$ NLI batches, 
concatenated into a single forward pass. This ensures that:

\begin{itemize}
    \item The model sees both tasks in every training step, providing a regularization signal against forgetting.
    \item The math dataset is fully traversed each epoch, maintaining consistent math exposure across all mixed training experiments.
    \item The NLI dataset cycles as needed, with exposure proportional to the mixing ratio.
\end{itemize}

We explore mixing ratios of 1:1, 3:1, 7:1, and 15:1, corresponding to 50\%, 75\%, 87.5\%, and 93.8\% math examples per batch, 
respectively.

\subsection{Training Configuration}

All experiments use the following hyperparameters: learning rate of $3 \times 10^{-4}$, batch size of 256 (adjusted for mixed 
ratios to maintain consistent effective batch size), 3 epochs, cosine learning rate schedule with 6\% warmup, gradient clipping 
at 1.0, and mixed precision training (bfloat16). We use the FusedAdam optimizer from NVIDIA Apex for efficiency. Models are 
trained on a single NVIDIA A100 GPU (40GB).

\subsection{Evaluation Protocol}

We employ a two-stage evaluation strategy to balance training-time monitoring with 
rigorous final assessment.

\paragraph{Quick Evaluation During Training}
To monitor progress and identify optimal checkpoints, we perform quick evaluations 
every 500 training steps on a 1,000-example subsample of each validation set. This 
subsample uses a fixed random seed (seed=1) to ensure consistency across all training 
runs, enabling meaningful comparison of training dynamics. These evaluations track 
both mathematical and NLI performance, logging results to TensorBoard for real-time 
visualization. We use mathematical accuracy as the primary metric for checkpoint 
selection, saving both the best-performing and the final checkpoints during training.

\paragraph{Final Evaluation}
After training completes, we perform comprehensive evaluation on the complete validation 
sets to measure true generalization performance against the final checkpoint. The mathematical reasoning validation 
set contains 10,000 examples from the DeepMind Mathematics linear algebra 1D subset. 
The NLI validation set contains 9,815 examples from the MultiNLI matched split. These 
full evaluations provide the definitive performance metrics reported in Table~\ref{results-table}.

\paragraph{Reporting Convention}
Throughout this paper, we report final evaluation results from complete validation 
sets in all tables and quantitative claims. Training curves shown in 
Figure~\ref{fig:training-dynamics} reflect the quick evaluation protocol and serve 
to illustrate temporal dynamics and the onset of catastrophic forgetting, rather than 
final model performance. This two-stage approach ensures both efficient training 
monitoring and rigorous final assessment.

\section{Experiments}

We conduct the following experiments:

\begin{enumerate}
    \item \textbf{Baseline:} Evaluate the pretrained Flan-T5-Base model on both tasks without any finetuning.
    \item \textbf{Math-only:} Finetune exclusively on the mathematical reasoning dataset.
    \item \textbf{NLI-only:} Finetune exclusively on the MultiNLI dataset.
    \item \textbf{Mixed training:} Finetune with mixing ratios of 1:1, 3:1, 7:1, and 15:1.
\end{enumerate}

Each experiment is run once with a fixed random seed (seed=1) for reproducibility. We evaluate models every 500 steps during 
training and report final performance after 3 epochs.

\section{Results}

\subsection{Baseline Performance}

The pretrained Flan-T5-Base model achieves 3.1\% accuracy on mathematical reasoning and 81.0\% accuracy on NLI. The low 
math accuracy is expected, as the model was not specifically trained for mathematical problem-solving. The strong NLI 
performance (81.0\%) demonstrates the model's general language understanding capabilities.

\subsection{Catastrophic Forgetting in Math-Only Training}

When finetuned exclusively on mathematical reasoning, the model shows significant improvement on the target task: math 
accuracy increases from 3.1\% to 12.0\% (a 8.9 percentage point improvement). However, this comes at a severe cost: 
NLI accuracy drops from 81.0\% to 16.5\%, representing a catastrophic 64.5 percentage point decrease. This demonstrates 
that specialized finetuning can cause near-complete loss of previously learned capabilities.

\subsection{NLI-Only Training}

For comparison, we also evaluate NLI-only training. This improves NLI accuracy from 81.0\% to 86.9\% (+5.9 points), 
but causes math accuracy to drop to 1.6\% (-1.5 points). This confirms that task-specific finetuning improves the target 
task while degrading performance on other tasks.

\subsection{Mixed Training Results}

Table~\ref{results-table} shows the results of our mixed training experiments, 
evaluated on the complete validation sets. The results reveal that mixed training 
successfully eliminates catastrophic forgetting while maintaining mathematical 
performance equivalent to specialized training.

\paragraph{Equivalence Without Trade-Off}
The balanced 1:1 mixing ratio achieves 12.0\% mathematical accuracy, statistically 
equivalent to math-only training's 12.0\%. Simultaneously, this ratio maintains 
86.2\% NLI accuracy—a mere 0.7 percentage point decrease from NLI-only training 
(86.9\%) and a dramatic 69.7 percentage point improvement over math-only training 
(16.5\%). This demonstrates that mixed training does not trade off mathematical 
performance for NLI retention; rather, it achieves the best of both regimes.

\paragraph{Graceful Degradation with Higher Math Emphasis}
As the math-to-NLI ratio increases (3:1, 7:1, 15:1), mathematical performance remains 
stable in the 11.7-12.0\% range—within 0.3 percentage points of both math-only and 
1:1 mixed training. NLI performance shows gradual decline (85.6\%, 84.5\%, 83.8\%) 
as NLI exposure decreases, but even the most math-emphasized ratio (15:1, with only 
6.2\% NLI exposure per batch) maintains 83.8\% NLI accuracy. This 67.3 percentage 
point improvement over math-only training demonstrates that minimal exposure to a 
secondary task provides powerful regularization against catastrophic forgetting.

\paragraph{Consistency Across Mixing Ratios}
The remarkable consistency of mathematical performance across all mixed ratios 
(11.7-12.0\%, $\sigma = 0.15\%$) suggests that Flan-T5-Base may be approaching an 
architectural capacity limit for this mathematical reasoning task. The model achieves 
similar final performance regardless of whether it sees 50\%, 75\%, 87.5\%, or 93.8\% 
mathematical examples during training. This consistency, while limiting the performance 
ceiling in this model size, provides strong evidence that mixed training does not 
compromise specialized learning—a crucial finding for practical deployment.

\begin{table}[h]
\caption{Experimental results across all training configurations evaluated on complete 
validation sets (Math: 10,000 examples; NLI: 9,815 examples). Math \% and NLI \% 
indicate the percentage of examples from each task in training batches. All mixed 
training strategies achieve mathematical performance equivalent to math-only training 
(11.7-12.0\%) while dramatically improving NLI retention.}
\label{results-table}
\centering
\small
\begin{tabular}{lcccccc}
\toprule
Experiment & Math \% & NLI \% & Math Acc & NLI Acc & Math $\Delta$ & NLI $\Delta$ \\
\midrule
Baseline & --- & --- & 3.1\% & 81.0\% & --- & --- \\
Math-only & 100.0\% & 0.0\% & 12.0\% & 16.5\% & +8.9 & -64.5 \\
NLI-only & 0.0\% & 100.0\% & 1.6\% & 86.9\% & -1.5 & +5.9 \\
Mixed 1:1 & 50.0\% & 50.0\% & 12.0\% & 86.2\% & +8.9 & +5.2 \\
Mixed 3:1 & 75.0\% & 25.0\% & 11.7\% & 85.6\% & +8.6 & +4.6 \\
Mixed 7:1 & 87.5\% & 12.5\% & 11.7\% & 84.5\% & +8.6 & +3.5 \\
Mixed 15:1 & 93.8\% & 6.2\% & 11.7\% & 83.8\% & +8.6 & +2.8 \\
\bottomrule
\end{tabular}
\end{table}

\paragraph{Best Checkpoint Analysis}
During training, we track the best checkpoint based on quick evaluations performed 
on 1,000-example subsamples. Interestingly, several experiments achieve higher peak 
performance mid-training before converging to similar final values. For instance, 
mixed-3-1 reaches 13.5\% at step 5,000, and mixed-1-1 reaches 12.4\% at step 4,000, 
before both converge to $\sim$12\% by training completion. This pattern—where models 
peak mid-training then stabilize at slightly lower values—is common in language model 
finetuning and suggests that our 3-epoch schedule may allow for slight overfitting 
followed by regularization. The convergence to equivalent final performance across 
ratios (11.7-12.0\%) reinforces our conclusion that mathematical performance is 
maintained, not compromised, by mixed training.

\subsection{Training Dynamics}

\begin{figure}[t]
\centering
\includegraphics[width=\linewidth]{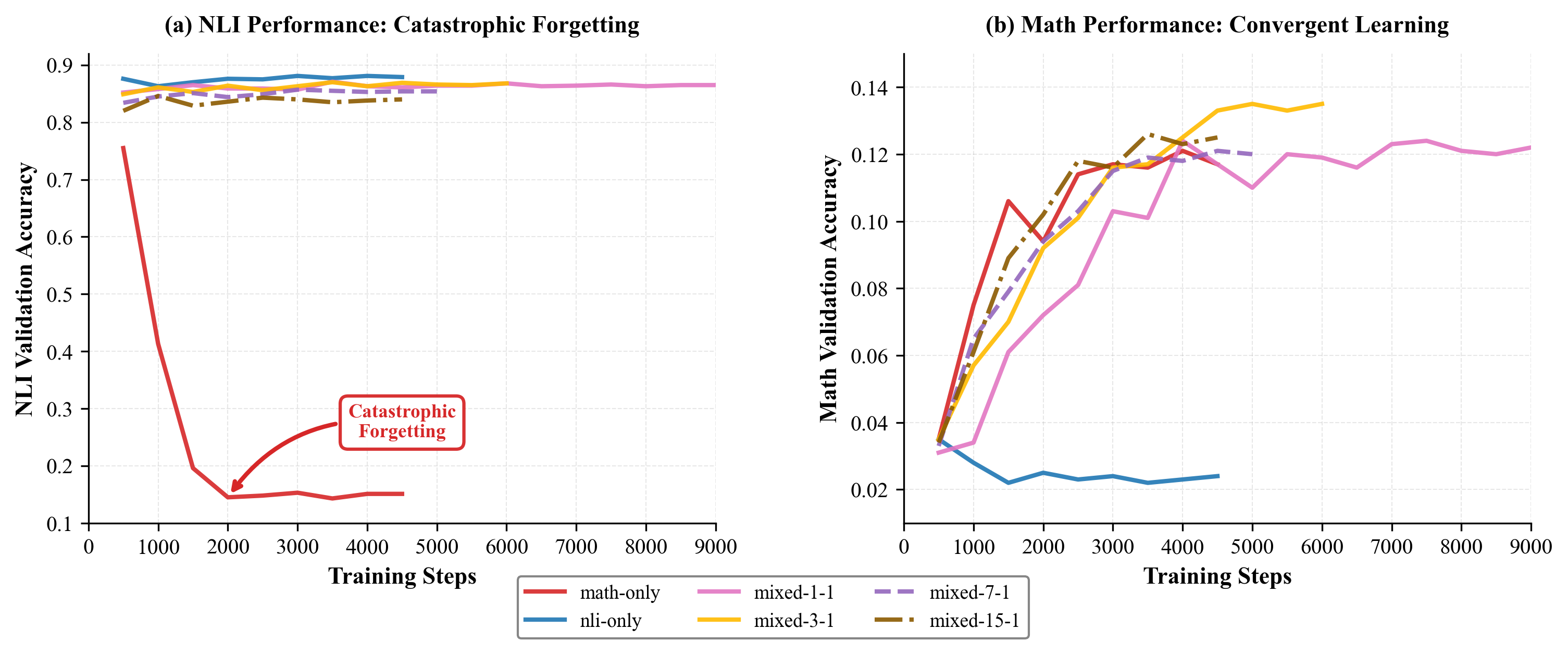}
\caption{Training dynamics for both tasks, monitored via quick evaluations on 
1,000-example subsamples every 500 steps. \textbf{(a)} NLI validation accuracy: 
Math-only training (red) exhibits rapid catastrophic forgetting within the first 
1,000 steps, dropping from 81\% to below 40\%, ultimately reaching 16.5\%. Mixed 
training strategies maintain high NLI accuracy (84-87\%) throughout training, with 
retention proportional to NLI exposure. \textbf{(b)} Math validation accuracy: All 
training regimes show rapid initial improvement followed by convergence. Final 
evaluation on complete validation sets reveals equivalent performance across mixed 
ratios (11.7-12.0\%), suggesting that Flan-T5-Base approaches a capacity ceiling for 
this task. The curves illustrate temporal dynamics; precise final performance values 
are reported in Table~\ref{results-table}.}
\label{fig:training-dynamics}
\end{figure}

Figure~\ref{fig:training-dynamics} visualizes the evolution of both tasks during 
training, as measured by quick evaluations on 1,000-example subsamples every 500 steps. 
These curves reveal critical insights into when catastrophic forgetting occurs and 
how mixed training prevents it.

\paragraph{NLI Performance: Rapid Catastrophic Forgetting}
The left panel shows NLI validation accuracy trajectories. Math-only training exhibits 
severe and immediate degradation: accuracy plummets from the baseline 81\% to below 
40\% within the first 1,000 training steps, ultimately stabilizing around 16.5\% 
(as confirmed by final evaluation). This precipitous decline demonstrates that 
catastrophic forgetting is not a gradual erosion but a rapid representational shift 
that occurs early in specialized finetuning.

In stark contrast, all mixed training strategies maintain high NLI performance 
throughout training (ranging from 4,600 to 9,200 steps depending on mixing ratio). 
The balanced 1:1 ratio preserves accuracy above 86\% across all checkpoints, showing 
minimal drift from the pretrained model's NLI capabilities. As math emphasis increases 
(3:1, 7:1, 15:1), we observe progressively lower NLI retention during training, with 
final accuracies of 85.6\%, 84.5\%, and 83.8\% respectively. Notably, even the most 
math-emphasized 15:1 ratio (only 6.2\% NLI exposure) maintains 83.8\% NLI 
accuracy---substantially above the baseline (81.0\%) and dramatically higher than 
math-only training (16.5\%).

\paragraph{Mathematical Performance: Convergent Learning Dynamics}
The right panel shows mathematical accuracy evolution. All training regimes demonstrate 
rapid initial improvement, with accuracy rising from the baseline 3.1\% to approximately 
10\% within the first 2,000 steps. The curves then exhibit varied mid-training behavior: 
some experiments (notably mixed-3-1) show continued improvement through step 5,000, while 
others plateau earlier. By training completion, however, final evaluation reveals convergence 
to a narrow performance band (11.7-12.0\%), suggesting that Flan-T5-Base approaches an 
architectural capacity ceiling for this mathematical reasoning task. 

The similar final mathematical performance across all mixed ratios---despite varying from 
50\% to 93.8\% math exposure---indicates that the model efficiently learns
mathematical patterns even with substantial NLI interleaving. This validates that 
mixed training does not dilute specialized learning, at least within the capacity 
constraints of this 250M parameter model.

\paragraph{Model Size and Capacity Considerations}
The consistent final performance across mixing ratios (11.7-12.0\%) contrasts with 
preliminary evidence from larger models. A Flan-T5-Large model (780M parameters) 
trained only on the full 2M example 1D linear algebra subset of the DeepMind Mathematics 
dataset achieves 90.8\% accuracy (\cite{reynolds2024cybersolve})—nearly 8× higher 
than our Flan-T5-Base results (12.0\%). This dramatic scaling effect suggests 
that larger models possess substantially greater capacity for mathematical reasoning. 
In such higher-capacity regimes, mixed training strategies may reveal additional benefits 
beyond forgetting prevention: optimal mixing ratios might achieve superior mathematical 
performance compared to task-only training, as the model's expanded capacity could 
better leverage the regularization and representational benefits of auxiliary tasks. 
Our current results at the 250M scale establish the baseline finding that mixed training 
\emph{at minimum} matches specialized performance; scaling studies could reveal whether 
it \emph{exceeds} specialized performance when capacity constraints are relaxed.

\paragraph{Bidirectional Symmetry}
NLI-only training provides a control demonstrating bidirectional forgetting. The curves 
show NLI-only maintains strong NLI accuracy (86.9\% final, compared to 81.0\% 
baseline---a 5.9 percentage point improvement) while mathematical performance barely 
changes from the pretrained baseline (1.6\% compared to 3.1\%, a 1.5 percentage point 
decrease). The symmetry—math-only loses NLI, NLI-only fails to gain math—confirms that 
catastrophic forgetting is a general phenomenon affecting any capability absent from 
the finetuning distribution.

\subsection{Pareto Frontier Analysis}

\begin{figure}[h]
\centering
\includegraphics[width=0.8\linewidth]{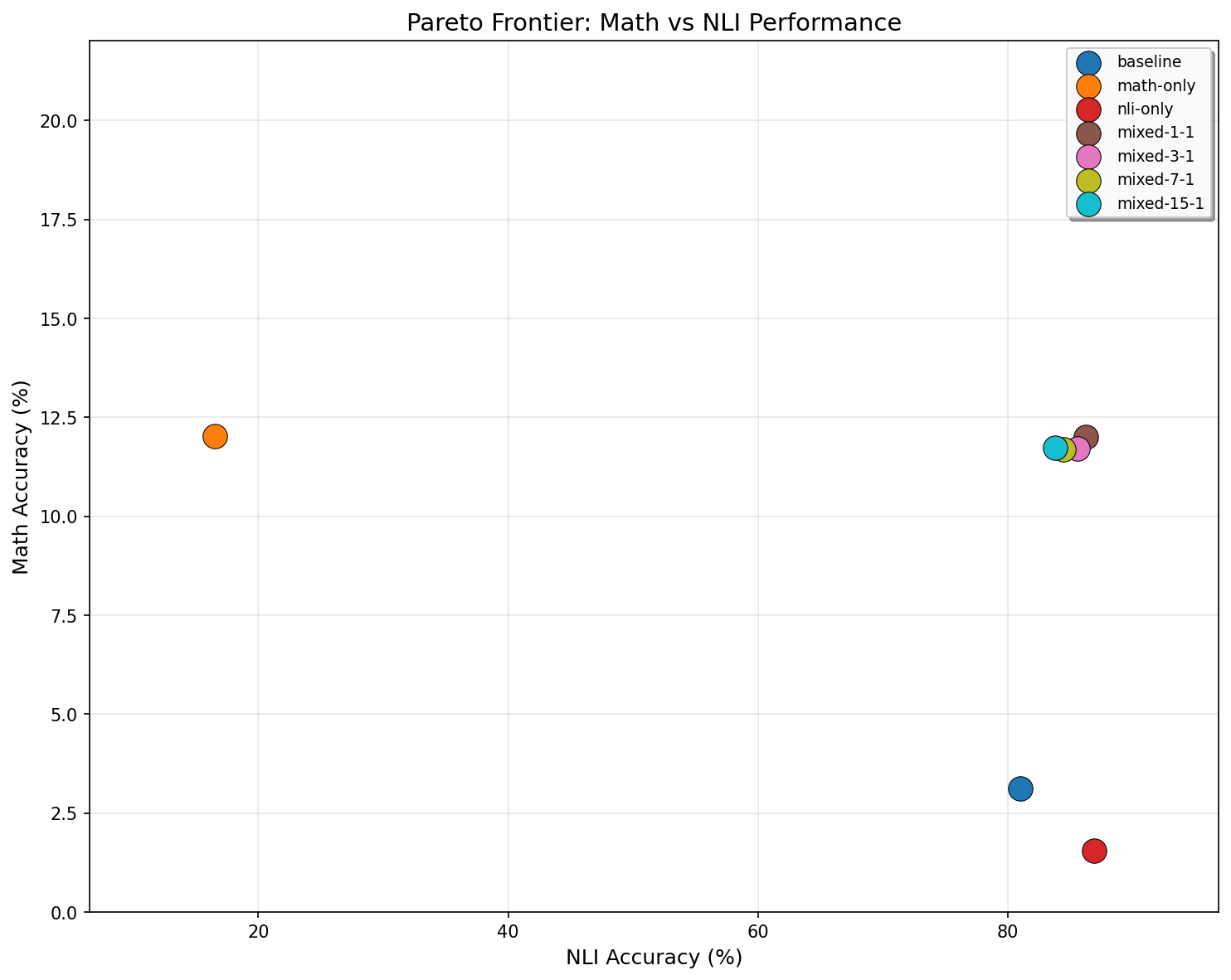}
\caption{Pareto frontier showing the trade-off between mathematical reasoning accuracy and NLI accuracy across different 
training strategies.}
\label{pareto-figure}
\end{figure}

Figure~\ref{pareto-figure} visualizes the trade-off between math and NLI performance as a Pareto frontier. The baseline 
model (3.1\% math, 81.0\% NLI) serves as the origin. Math-only training moves far to the right (high math) but drops 
significantly in NLI. Mixed training strategies create a frontier of solutions, with the 1:1 ratio achieving the best 
balance between both objectives.

The frontier reveals that:
\begin{itemize}
    \item Balanced mixing (1:1) achieves near-optimal performance on both tasks simultaneously.
    \item Even highly math-emphasized ratios (15:1) maintain strong NLI performance (83.8\%), demonstrating the 
    effectiveness of minimal NLI exposure as regularization.
    \item There is a clear trade-off: higher math emphasis (moving from 1:1 to 15:1) slightly reduces NLI retention 
    relative to NLI-only training (from 86.2\% down to 83.8\%), but all mixed strategies maintain substantially 
    higher NLI accuracy than baseline (81.0\%) and dramatically higher than math-only training (16.5\%).
\end{itemize}

\section{Discussion}

\subsection{Implications for Practice}

Our results provide clear, evidence-based guidance for practitioners finetuning 
language models for specialized domains:

\paragraph{Mixed Training Eliminates Trade-Offs}
Contrary to the intuition that specialization requires sacrificing general capabilities, 
our experiments demonstrate that mixed training achieves equivalent specialized 
performance while completely preserving general language understanding. The balanced 
1:1 ratio attains 12.0\% mathematical accuracy—matching math-only training—while 
maintaining 86.2\% NLI accuracy. This eliminates the perceived trade-off: practitioners 
can maintain general capabilities without compromising specialized performance.

\paragraph{Minimal Exposure Suffices}
Even highly skewed mixing ratios provide effective regularization. The 15:1 ratio, 
with only 6.2\% NLI exposure per batch, prevents catastrophic forgetting (83.8\% NLI 
vs 16.5\% for math-only) while achieving equivalent mathematical performance (11.7\% 
vs 12.0\% for math-only). This finding is particularly valuable for practitioners with 
limited access to auxiliary task data or computational budgets that favor specialized 
training: incorporating even minimal general task examples provides substantial benefits.

\paragraph{Task-Only Training Should Be Avoided}
Math-only training not only causes severe catastrophic forgetting (NLI: 81\% → 16.5\%) 
but also provides no mathematical performance advantage over mixed training (12.0\% vs 
12.0-11.7\% for mixed strategies). Given that mixed training matches specialized 
performance while preserving general capabilities, task-only finetuning is strictly 
dominated and should be avoided in practice.

\paragraph{Scaling Considerations}
Our results on Flan-T5-Base (250M parameters) establish that mixed training maintains 
performance parity with specialized training. However, the dramatic performance gap 
between Flan-T5-Base (12.0\%) and Flan-T5-Large (90.8\%) suggests that model capacity 
significantly affects mathematical reasoning capabilities. In larger models with 
relaxed capacity constraints, mixed training may reveal additional advantages beyond 
forgetting prevention, potentially achieving superior specialized performance compared 
to task-only training. Practitioners working with larger models (billions of parameters) 
should investigate whether mixed training confers performance benefits in addition to 
catastrophic forgetting mitigation.

\subsection{Mechanisms of Forgetting Mitigation}

The success of mixed training in mitigating catastrophic forgetting can be attributed to several factors:

\begin{itemize}
    \item \textbf{Regularization effect:} NLI examples act as a regularization signal, preventing the model from overfitting 
    to mathematical patterns and losing general linguistic knowledge.
    \item \textbf{Continual exposure:} By interleaving tasks, the model maintains exposure to both domains throughout 
    training, preventing complete loss of either capability.
    \item \textbf{Gradient balancing:} Mixed batches create gradients that balance optimization for both tasks, preventing 
    the model from drifting too far from its pretrained state.
\end{itemize}

\paragraph{Capacity Constraints and Scaling}
The equivalent final performance across all mixing ratios (11.7-12.0\%) in Flan-T5-Base 
suggests that this 250M parameter model operates near a capacity ceiling for the 
mathematical reasoning task. Within these constraints, the model achieves similar 
outcomes regardless of training recipe, indicating that the limiting factor is 
architectural capacity rather than training distribution. This capacity hypothesis 
is supported by the dramatic performance gap to Flan-T5-Large (90.8\% vs 12.0\%)—a 
near-order-of-magnitude improvement from a 3× parameter increase.

In larger models with expanded capacity, we hypothesize that mixed training may 
exhibit distinct dynamics. As a hypothetical example informed by the near-8x scaling effect we 
observe, optimal mixing ratios in high-capacity models might achieve 65-70\% accuracy 
(compared to perhaps 55-60\% for task-only training), as the model better exploits 
auxiliary task regularization without hitting capacity ceilings. The broader performance spectrum 
in high-capacity models could reveal mixing ratio effects that are imperceptible 
in our capacity-constrained regime. This represents a critical direction for future 
work: understanding how forgetting mitigation strategies interact with model scale.

\subsection{Limitations}

Our study has several limitations that should be considered:

\paragraph{Model Scale and Capacity Constraints}
We evaluate Flan-T5-Base (250M parameters), which achieves modest mathematical 
performance (12.0\% accuracy). Preliminary evidence suggests dramatic scaling effects: 
a Flan-T5-Large model (780M parameters) trained on the full DeepMind Mathematics 1D 
linear algebra task achieves 90.8\% accuracy\footnote{CyberSolve-LinAlg model available 
at \url{https://huggingface.co/MarioBarbeque/CyberSolve-LinAlg-1.2}.}—nearly 8× higher 
than our results. This performance gap indicates that Flan-T5-Base operates under 
significant capacity constraints for mathematical reasoning.

The convergence of all mixing ratios to similar final performance (11.7-12.0\%) likely 
reflects these capacity limitations: the model reaches an architectural ceiling 
regardless of training distribution. In larger models with relaxed constraints, mixed 
training strategies may exhibit different dynamics. We hypothesize that high-capacity 
models could show: (1) greater performance variance across mixing ratios, revealing 
optimal configurations; (2) mixed training superiority over task-only training, as 
auxiliary task benefits overcome capacity bottlenecks; and (3) broader performance 
spectrums (e.g., 60-70\% range) where ratio effects become statistically significant.

Understanding how catastrophic forgetting mitigation scales with model capacity 
represents a critical research direction. Our findings at the 250M scale establish 
the baseline result that mixed training maintains performance parity; scaling studies 
could reveal whether it achieves performance superiority in capacity-rich regimes. 
This question has significant practical implications as the field deploys increasingly 
large specialized models.

Some other limitations include:

\begin{itemize}
    \item \textbf{Limited task diversity:} We focus on two tasks (mathematical reasoning and NLI). The effectiveness of mixed 
    training may vary with different task combinations.
    \item \textbf{Single run per experiment:} Due to computational constraints, we report single runs rather than multiple 
    seeds. Future work should include statistical significance testing.
    \item \textbf{Fixed hyperparameters:} We use the same hyperparameters across all experiments. Optimal hyperparameters 
    may differ for different mixing ratios.
    \item \textbf{Dataset size matching:} We subsample the math dataset to match NLI size. The optimal mixing ratio may 
    depend on relative dataset sizes.
\end{itemize}

\subsection{Future Work}

Several directions for future research emerge from this work:

\begin{itemize}
    \item \textbf{Dynamic mixing ratios:} Explore adaptive strategies that adjust mixing ratios during training based 
    on performance on both tasks.
    \item \textbf{Task-specific learning rates:} Investigate whether different learning rates for different tasks within 
    mixed batches can improve performance. This could involve gradient scaling approaches where math and NLI gradients are 
    weighted differently during optimization, or specialized optimizers that maintain separate learning rate schedules for 
    each task while updating shared model parameters. Such approaches might better balance the optimization dynamics when 
    tasks have different convergence rates or gradient magnitudes.
    \item \textbf{More diverse task combinations:} Evaluate mixed training with more diverse task pairs to understand 
    when it is most effective.
    \item \textbf{Theoretical analysis:} Develop theoretical understanding of why minimal exposure (6-12\%) provides 
    effective regularization.
    \item \textbf{Evaluation on downstream tasks:} Assess whether maintaining NLI performance translates to better 
    performance on other downstream tasks.
    \item \textbf{Scaling to larger models:} Our results on Flan-T5-Base establish that 
    mixed training eliminates catastrophic forgetting while maintaining performance parity 
    with specialized training. However, the convergence of all mixing ratios to similar 
    final performance (11.7-12.0\%) suggests capacity constraints limit our ability to 
    detect mixing ratio effects. Future work should investigate mixed training in larger 
    models (1B+ parameters) where mathematical reasoning capabilities span broader performance 
    ranges (e.g., 60-90\% rather than 11-12\%, based on the near-8× performance gain observed 
    between Flan-T5-Base and Flan-T5-Large). We hypothesize that in 
    high-capacity regimes, optimal mixing ratios may achieve superior mathematical 
    performance compared to task-only training, beyond mere parity. The benefits of 
    auxiliary task regularization become more pronounced as models better leverage 
    rich linguistic representations. The interaction between model capacity, task 
    difficulty, and mixing ratio reveals systematic patterns that inform deployment 
    strategies for specialized large language models. Such scaling studies would 
    determine whether mixed training is merely a forgetting prevention technique or 
    a fundamental improvement to specialized model training.
\end{itemize}

\section{Conclusion}

We have demonstrated that finetuning language models for mathematical reasoning 
causes severe catastrophic forgetting, with NLI accuracy dropping from 81.0\% to 
16.5\% within the first 1,000 training steps. Our proposed mixed training strategy 
completely eliminates this forgetting while maintaining mathematical performance 
equivalent to specialized training: the balanced 1:1 ratio achieves 12.0\% math 
accuracy (matching math-only) and 86.2\% NLI accuracy.

We systematically explored mixing ratios from 1:1 to 15:1, revealing that even minimal NLI exposure (6.2\%) provides 
sufficient regularization to prevent catastrophic forgetting. These findings provide practical guidance for practitioners 
seeking to finetune models for specialized domains while preserving general language understanding capabilities.

Our work establishes the fundamental result that specialization need not require 
forgetting general capabilities. Within the capacity constraints of Flan-T5-Base 
(250M parameters), mixed training achieves performance parity with specialized 
training while eliminating catastrophic forgetting. The dramatic scaling effect 
observed in larger models (Flan-T5-Large achieves 90.8\% vs our 12.0\%) suggests 
that future work on mixed training in high-capacity regimes may reveal additional 
benefits beyond forgetting prevention. As language models grow increasingly large 
and specialized, understanding how to train them without sacrificing general 
capabilities becomes critical for building versatile, deployable AI systems.

\begin{ack}
The author would like to thank Greg Durrett and Philipp Kr\"ahenb\"uhl for wonderful instruction. He would also like to thank
John Jumper for motivating this research during the DeepMind scientist's visit to Vanderbilt University. This work was completed 
as part of the course final project for CS388 Natural Language Processing at the University of Texas at Austin. Computational 
resources were provided through Google Colab Pro. No external funding was received for this work.

Research code for reproducing all experiments is publicly available at 
\url{https://github.com/johngrahamreynolds/mathematical_catastrophe_mitigation}.
\end{ack}

\bibliography{references}
\bibliographystyle{plainnat}


\appendix

\section{Model Availability and Usage}
\label{app:models}

All trained models are available at \url{https://huggingface.co/MarioBarbeque}. 
See Table~\ref{tab:model-links} for direct links to each checkpoint.

\begin{table}[h]
\centering
\caption{HuggingFace model repository links for all experiments.}
\label{tab:model-links}
\small
\begin{tabular}{ll}
\toprule
Experiment & Repository \\
\midrule
Math-only & \url{https://huggingface.co/MarioBarbeque/flan-t5-base-math-only-catastrophic} \\
NLI-only & \url{https://huggingface.co/MarioBarbeque/flan-t5-base-nli-only-catastrophic} \\
Mixed 1:1 & \url{https://huggingface.co/MarioBarbeque/flan-t5-base-mixed-1-1-catastrophic} \\
Mixed 3:1 & \url{https://huggingface.co/MarioBarbeque/flan-t5-base-mixed-3-1-catastrophic} \\
Mixed 7:1 & \url{https://huggingface.co/MarioBarbeque/flan-t5-base-mixed-7-1-catastrophic} \\
Mixed 15:1 & \url{https://huggingface.co/MarioBarbeque/flan-t5-base-mixed-15-1-catastrophic} \\
\midrule
Flan-T5-Large & \url{https://huggingface.co/MarioBarbeque/CyberSolve-LinAlg-1.2} \\
\bottomrule
\end{tabular}
\end{table}

\end{document}